\documentclass[runningheads]{llncs}
\usepackage[T1]{fontenc}
\usepackage{graphicx} % For including images
\usepackage{float}    % For the [H] placement specifier
\usepackage{hyperref}
\usepackage{booktabs}
\usepackage{subcaption}  % For subfigures

\usepackage{pifont}
\usepackage{newunicodechar} % Maps Unicode to commands
\usepackage{comment}
\usepackage{siunitx}

\newunicodechar{✓}{\ding{51}}   % Checkmark
\newunicodechar{✗}{\ding{55}}   % Ballot X (Heavy Ballot X)

\begin{document}

%% ---------------------------------------------------------------
%% ICDAR Full Paper: max 17 pages including figures and references
%% ---------------------------------------------------------------

%
\title{ICDAR 2026 Competition on Information Extraction from Atomic Layer Deposition/Etching (ALD/E) Scientific Figures}

\titlerunning{ICDAR 2026 Sci-ImageMiner Competition}

\author{
Fahad Ahmed\inst{1}\orcidID{0000-0001-9201-1580} \and
% Jennifer D'Souza\inst{1}\orcidID{0000-0002-6616-9509} \and
Sören Auer\inst{1,2}\orcidID{0000-0002-0698-2864} \and
Jennifer D'Souza\inst{1}\orcidID{0000-0002-6616-9509}
}
\authorrunning{F. Ahmed et al.}

\institute{
TIB - Leibniz Information Centre for Science and Technology, Hannover, Germany \\
\email{\{Fahad.Ahmed,Auer,Jennifer.DSouza\}@tib.eu}
\and
L3S Research Center, Leibniz University, Hannover, Germany \\
\email{auer@l3s.de}
}

%
\maketitle              % typeset the header of the contribution
%

% ---- ABSTRACT ----
\begin{abstract}
%The abstract should briefly summarize the contents of the paper in 150--250 words.
Scientific figure comprehension and reasoning using multimodal AI requires integrating visual perception with domain-specific reasoning to extract meaningful knowledge, often not presented in the text of a research publication.
% We introduce *Sci-ImageMiner*, a large-scale, expert-annotated benchmark dataset for scientific figures in ALD/E research, providing a standardized framework for evaluating end-to-end multimodal understanding.
The Sci-ImageMiner benchmark dataset, accompanied by a community-driven competition, 
% across four end-to-end complementary tasks, 
raises the bar over prior scientific competitions by curating a comprehensive, expert-annotated dataset across four end-to-end complementary tasks. 
% , we engineered a hard competition dataset and proposed the recent
The competition attracted 68 active participants and 1,263 public/private submissions from 9th January 2026 to 8th April 2026.
Our results show that state-of-the-art multimodal models perform well on classification and summarization tasks but struggle with data extraction and scientific reasoning, particularly in visual question-answering. These findings reveal key limitations and highlight challenges and opportunities for improving domain-aware multimodal AI systems. Overall, the Sci-ImageMiner benchmark and competition establish a rigorous platform for advancing research in scientific figure comprehension and reasoning and demonstrate the potential of state-of-the-art approaches for a challenging and complex research area.

\end{abstract}
%
%
%

% ---- INTRODUCTION ----
\section{Introduction}
\label{sec:introduction}

%% 1. What
% Scientific figure comprehension and reasoning are increasingly significant challenges in multimodal AI, particularly in domains such as materials science and chemistry where critical knowledge is often conveyed through complex visual artifacts rather than text alone. In Atomic Layer Deposition and Etching (ALD/E) research, figures especially quantitative plots encode essential quantitative and procedural insights that require domain-specific reasoning to interpret. 

%% 2. Why
% Existing benchmarks and prior ICDAR competitions on scientific figure understanding \cite{chen2024icdar,yang2017icdar2017,kayal2021icdar} have made significant progress on a limited set of tasks, yet they do not fully capture the complexity and necessity of real-world scientific figure comprehension across diverse complementary tasks.
% In particular, most available datasets focus on generic or synthetic charts and address isolated tasks such as classification or partial data extraction, rather than supporting end-to-end multimodal reasoning across diverse complementary tasks including classification, structured data extraction, summarization, and visual question-answering. Furthermore, while recent advances in AI have enabled progress in multimodal understanding and scientific processes, current state-of-the-art remain limited in handling specialized scientific figures due to their complexity and the lack of expert-curated, domain-grounded benchmarks.

%% -- What?
Scientific figure comprehension and reasoning using multimodal AI is the ability to interpret, analyze, and derive meaningful scientific knowledge from figures in research publications.
% Unlike general image understanding, 
In particular, figures such as charts or plots, not only require visual perception,
% e.g. axes, curves, points and lines
but also domain-specific knowledge, e.g., interpreting trends, relationships, etc., for reasoning.
% These figures embed complex information, results and insights, which are often not explicitly mentioned in the text.
% and is significant in scientific communication and knowledge extraction.
% Scientific figure comprehension and reasoning
% It can be described as an end-to-end multimodal information extraction systematic process,
% which can be divided into several complementary tasks such as classification, data table extraction, summarization and visual question-answering.
% Progressively moving from low-level visual parsing to high-level scientific reasoning.
This helps to capture a deeper level of scientific comprehension in an 
% progressive 
end-to-end workflow which enables the transformation of visual evidence into structured knowledge and actionable insights~\cite{wanselin2022analysing}.
%% -- Why: ALD/E in Material Science
Moreover, in semiconductor manufacturing, Atomic layer deposition (ALD) and Atomic layer etching (ALE) are 
% cornerstone
foundational technologies in materials science for advanced electronic and functional materials applications. and next generation nanoelectronics at atomic-scale~\cite{george2010atomic,kanarik2015overview}.

%% -- Why?
%% -- Why: Current Multimodal Models
Recent research using multimodal AI shows outstanding performance on natural images while struggling with complex semantics and abstract representation of scientific figures, essentially due to the lack of domain-specific datasets and a need for domain-specific reasoning capabilities~\cite{li2024multimodal,jablonka2025vision}.
% Studies show that current multimodal models perform exceptionally well on basic perception tasks but often fail in multi-step scientific reasoning process, 
% which identifies a significant gap between visual perception and true scientific reasoning
% \cite{jablonka2025vision}.
%% -- Why: Previous ICDAR Competitions 
% Furthermore, prior 
Previous ICDAR competitions~\cite{chen2024icdar,yang2017icdar2017} on scientific figure understanding despite 
significant contributions, 
% contributed significantly, however, 
% do not 
lack the ability to fully capture the complexity and necessity of  
end-to-end real-world scientific figure 
% understanding.
comprehension and reasoning.
% across progressive complementary tasks.
%% -- Why: Existing general-purpose datasets
General-purpose chart 
% comprehension 
datasets either focus on real-world \cite{masry-etal-2022-chartqa}
% kantharaj-etal-2022-chart-to-text
or synthetic \cite{methani2020plotqa} data,
% xia2023structchart-simchart9k
%% -- Why: Existing Scientific figure comprehension and reasoning datasets in Material Science
% Scientific figure comprehension and reasoning 
while datasets~\cite{alampara2024macbench,zhang2025matscibench} in material science remain limited in scope, 
% , often focusing on a narrow
% and 
range of quantitative figures,
% visualizations 
and task formulations.
%% -- Why: The gap and the need
These limitations 
% of existing datasets and competitions, 
% ranging from limited chart diversity, reliance on synthetic data, narrow task coverage and limited domain specificity,
underscore a critical gap, 
% in current scientific figure comprehension and reasoning.
in particular, there remains a lack of a large-scale, expert-curated dataset that supports end-to-end multimodal reasoning  
% across multiple complementary tasks 
on authentic, domain-grounded scientific figures.
% Collectively, these limitations highlight the need for a comprehensive benchmarks that integrate diverse figure types and support complementary tasks using a progressive end-to-end pipeline, for robust scientific figure comprehension and reasoning in a highly specialized scientific domain.
%  Furthermore, while recent advances in AI have enabled progress in multimodal understanding and scientific processes, current state-of-the-art remain limited in handling specialized scientific figures due to their complexity and the lack of expert-curated, domain-grounded benchmarks.

%% 3. How
% To bridge this gap, we introduce Sci-ImageMiner, a comprehensive benchmark dataset specifically designed for Atomic Layer Deposition and Etching (ALD/E) research in material science. By incorporating diverse real-world scientific figures and supporting a unified framework spanning classification, structured data extraction, summarization, and visual question-answering, Sci-ImageMiner provides a more realistic and challenging benchmark for advancing multimodal scientific comprehension and reasoning in a highly specialized scientific domain.

%% -- How
To address these limitations, we introduced Sci-ImageMiner, 
% built upon 
the first comprehensive, expert-annotated, domain-specific dataset of ALD/E scientific figures.
% research.
% It provides a standardized, 
% large-scale, 
% expert-annotated benchmark 
% With a unified evaluation protocol, enabling systematic assessment 
% of multimodal AI approaches 
% on authentic, figure-intensive scientific data.
% on end-to-end scientific comprehension and reasoning pipeline.
Through this competition, we hope to address several key questions:  
% regarding the current state of multimodal AI models in scientific figure comprehension and reasoning:
\textbf{RQ1:} How well do current multimodal models understand and reason over domain-specific scientific figures?
% on particularly in 
% in a highly-specialized domain i.e.
% ALD/E in material science?
\textbf{RQ2:} Which of the approaches
% , techniques and strategies 
significantly improve the 
capabilities of multimodal models?
% for scientific figure comprehension and reasoning?
\textbf{RQ3:} Which of the competition tasks are most challenging even for state-of-the-art multimodal models?
% even after fine-tuning models and applying various techniques and approaches?
By engaging the research community through this competition, we aim to investigate 
% key questions regarding the current capabilities and limitations of multimodal AI models in specialized scientific domains, 
% and identify 
effective 
% modeling 
strategies 
% that improve performance, 
and analyze which aspects of scientific figure comprehension remain most challenging,  
% Through this initiative, 
% The competition not only establishes a 
via a rigorous and realistic platform, 
% but also 
fostering the development of robust, domain-aware multimodal AI systems capable of advancing scientific knowledge comprehension and reasoning. The competition dataset and evaluation scripts are available under a CC BY 4.0 license~\footnote{\label{sci-imageminer-repo}\url{https://github.com/sciknoworg/sci-imageminer/}}.
% ~\footnote{\url{https://github.com/sciknoworg/ALD-E-ImageMiner/tree/main/icdar2026-competition-data}}.

% -- Paper structure
The remainder of this paper is organized as follows. Section~\ref{sec:dataset} introduces the dataset curation 
% including its construction, organization,
and annotation process. Section~\ref{sec:competition} presents the 
% ICDAR 2026 
competition 
% detailing its 
% structure,
organization, 
timeline, tasks, and evaluation protocols.
% , and evaluation protocols. 
Section~\ref{sec:results_and_discussions} reports baseline and competition results,  summarizing participant approaches and a comparative analysis.
% across all four tasks.
% , including an analysis of the top-performing approaches and comparative insights into different methodologies.
Finally, Section~\ref{sec:conclusions} concludes the paper with 
% a discussion of 
key findings, limitations, future research directions, and acknowledgments.

\section{Sci-ImageMiner Benchmark Dataset}
\label{sec:dataset}

% In this section, we present the Sci-ImageMiner benchmark dataset. In Section~\ref{sec:dataset_collection_preperation}, we discuss the collection and preparation of data. Section~\ref{sec:dataset_organization} describes how the dataset is organized in a structured hierarchy and how it is distributed into standard splits. Finally, Section~\ref{sec:dataset_annotations} discuss the end-to-end annotation process in detail.
% ~\footnote{\url{https://github.com/opendatalab/MinerU}}\cite{wang2024mineruopensourcesolutionprecise}
%%
\subsection{Data Collection \& Preparation}
\label{sec:dataset_collection_preperation}

We collected 205 research publications in ALD/E materials science,
% with a specific focus on ALD/E, 
encompassing both experimental and simulation-based studies. 
% The dataset includes 66 experimental and 58 simulation studies related to ALD, as well as 46 experimental and 35 simulation studies focused on ALE.
% Identifying 49 figure types,
% A 49-figures taxonomy was developed with a \href{https://github.com/sciknoworg/ALD-E-ImageMiner/blob/main/figure_taxonomy.png}{visualization} and \href{https://github.com/sciknoworg/ALD-E-ImageMiner/blob/main/figure_taxonomy.tsv}{descriptions} shared in the \href{https://github.com/sciknoworg/ALD-E-ImageMiner}{Github repository}~\footnote{\url{https://github.com/sciknoworg/ALD-E-ImageMiner}}.
A 49-figures taxonomy was developed with a \href{https://github.com/sciknoworg/sci-imageminer/blob/main/figure_taxonomy.png}{visualization} and \href{https://github.com/sciknoworg/sci-imageminer/blob/main/figure_taxonomy.tsv}{descriptions} shared in the \href{https://github.com/sciknoworg/sci-imageminer/}{Github repository}~\textsuperscript{\ref{sci-imageminer-repo}}
% ~\footnote{\url{https://github.com/sciknoworg/sci-imageminer/}}.
% To systematically process these publications, we employed 
% We utilized 
% \href{https://github.com/opendatalab/MinerU}{MinerU}

% as illustrated below:
\begin{figure}[htb]
% \begin{figure}[H]
  \centering
  \includegraphics[width=0.7\linewidth]{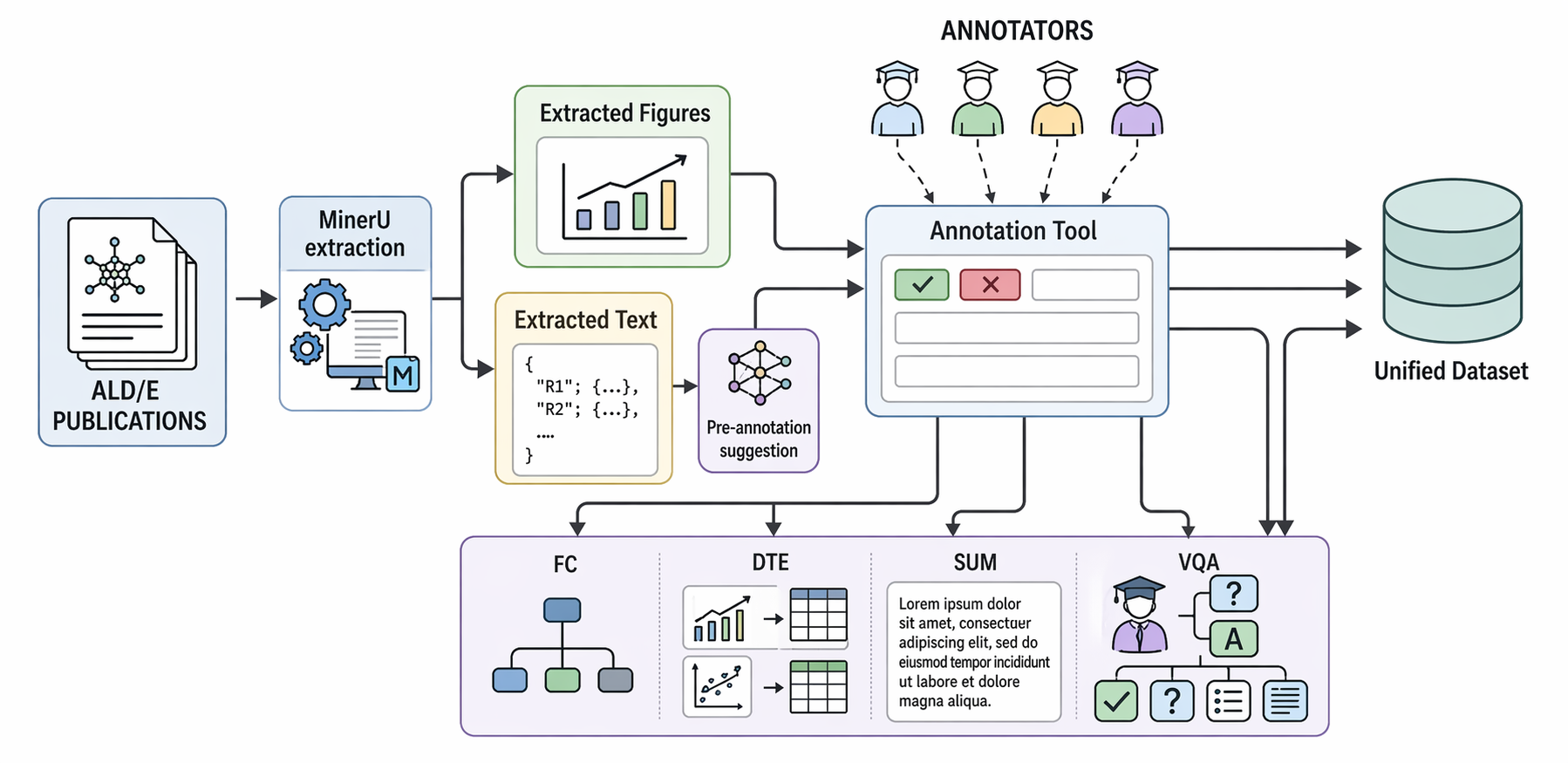}
  \caption{Overview of the Sci-ImageMiner benchmark end-to-end curation workflow.}
  \label{fig:annotation-workflow}
\end{figure}

MinerU~\cite{wang2024mineru}
% ~\cite{wang2024mineruopensourcesolutionprecise}, 
% which enables 
is utilized 
for the extraction of textual content in structured JSON format
% (i.e., sections, paragraphs, and figure captions)---
% as well as
along with high-resolution figures in JPEG format.
This workflow ensures the preservation of both the semantic structure and visual fidelity of the source documents for downstream multimodal analysis 
as illustrated in Figure~\ref{fig:annotation-workflow}.
% The dataset distribution and statistics are shown in Table~\ref{tab:dataset_stats}. 

% ---- ORGANIZATION ----
\subsection{Dataset Organization}
\label{sec:dataset_organization}

\begin{figure}[H]
  \centering
  \includegraphics[width=0.7\linewidth]{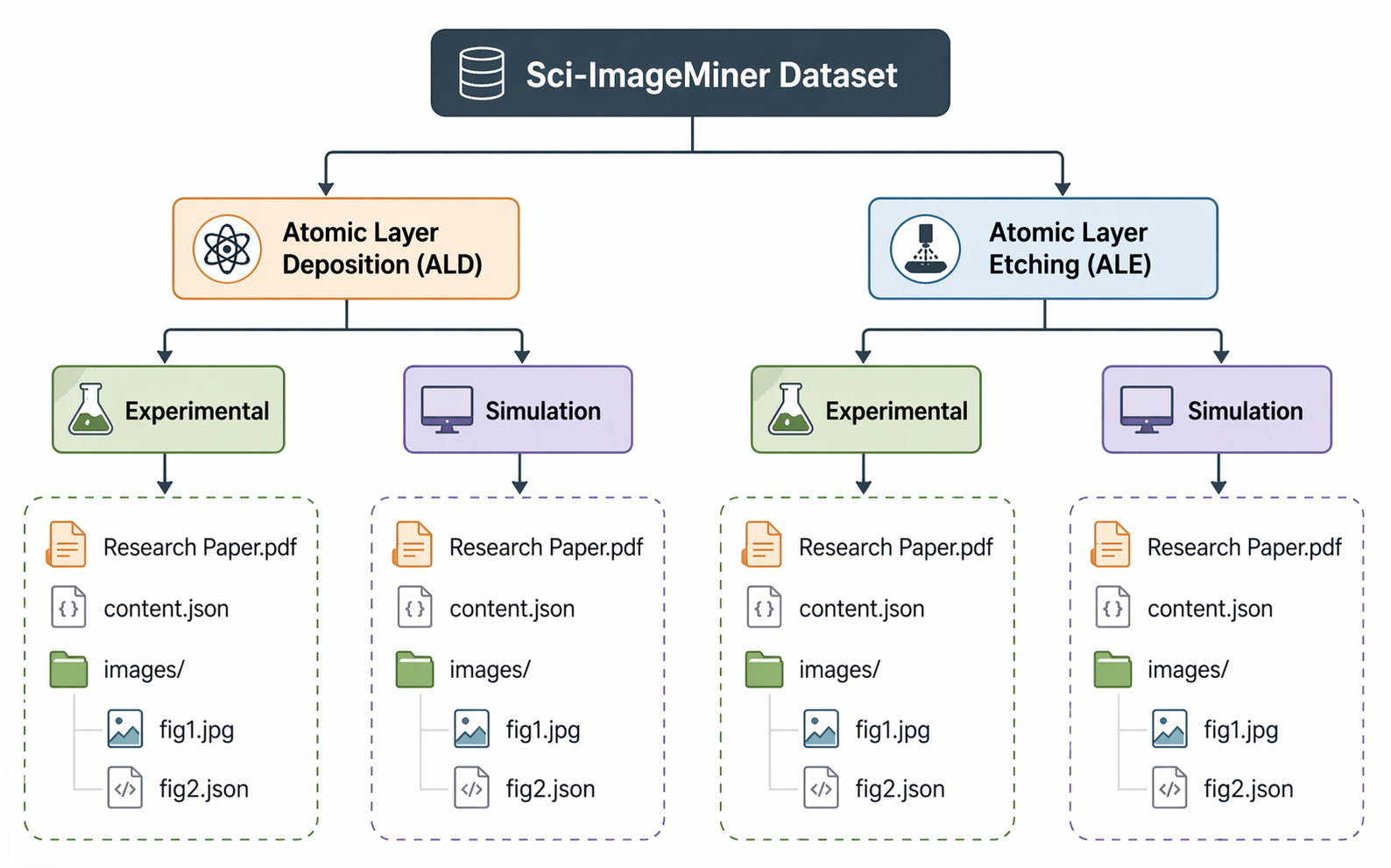}
  \caption{Illustration describing the Sci-ImageMiner dataset organization hierarchy. Research papers are numerically indexed, containing the PDF, a "content.json" (sections and figure captions), along with figures and annotation JSON files inside the "images" directory.}
  \label{fig:dataset_organization_hierarchy}
\end{figure}

The dataset is organized into two primary categories:
% corresponding to key surface-chemical techniques in materials science:
(i) ALD and (ii) ALE; which are further divided into: 
% two subcategories:
(i) experimental and (ii) simulation-based studies.
% Within each subcategory, 
% Individual 
% Research papers are represented as numerically indexed directories containing the original PDF, a `content.json` file with structured textual data (including sections and figure captions), and an `images` directory comprising extracted figures alongside their corresponding annotation JSON files.
% Each sample is uniquely identified using a hierarchical naming convention of the form main\_category/sub\_category/paper\_number/figure\_caption\_label.
The dataset is partitioned into standard 
% training, development, and test 
train/dev/test 
splits with the statistics shown in Table~\ref{tab:dataset_stats} and the organization as illustrated in Figure~\ref{fig:dataset_organization_hierarchy}.
% to facilitate systematic evaluation as illustrated in Figure~\ref{fig:dataset_organization_hierarchy}.
% Detailed statistics of the Sci-ImageMiner competition dataset are presented in Table~\ref{tab:dataset_stats}
% The Sci-ImageMiner dataset comprises a total of 1,951 figures extracted from 205 ALD/E research publications. Among these, 1,404 figures correspond to quantitative plots that have been meticulously annotated by domain experts to support a diverse range of scientific comprehension and reasoning tasks, including classification, structured data extraction, summarization, and visual question-answering.

% \begin{figure}[!htb]

%% TODO: Use the ALD/E distribution
\begin{table}[htb]
% \begin{table}[H]
\caption{Overview of the Sci-ImageMiner Dataset Statistics, showing distribution by each dataset split, ALD/ALE, and Experimental/Simulation studies. }
\centering
    \scriptsize
    \setlength{\tabcolsep}{3pt}
    \renewcommand{\arraystretch}{1.05}
    \resizebox{\columnwidth}{!}{%
    % \begin{tabular}{l l l l l l l l l l l}
    \begin{tabular}{l r r r r r r r r r r}
        \toprule
        % Split Name & atomic-layer-deposition & ~ & ~ & ~ & atomic-layer-etching & ~ & ~ & ~ & Total & ~ \\
        \textbf{Split Name} & \multicolumn{4}{c}{\textbf{ALD}} & \multicolumn{4}{c}{\textbf{ALE}} &  \multicolumn{2}{c}{\textbf{Total}} \\
        % ~ & experimental-usecase & ~ & simulation-usecase & ~ & experimental-usecase & ~ & simulation-usecase & ~ & ~ & ~ \\
        ~ & \multicolumn{2}{c}{\textbf{Experimental}} & \multicolumn{2}{c}{\textbf{Simulation}} & \multicolumn{2}{c}{\textbf{Experimental}} & \multicolumn{2}{c}{\textbf{Simulation}} \\
        ~ & Papers & Figures & Papers & Figures & Papers & Figures & Papers & Figures & Papers & Figures \\
        \midrule
        \textit{Train} & 42 & 330 & 36 & 350 & 28 & 257 & 22 & 243 & 128 & 1,180 \\
        \textit{Dev} & 6 & 50 & 6 & 59 & 5 & 55 & 3 & 37 & 20 & 201 \\
        \textit{Test} & 18 & 172 & 16 & 148 & 13 & 129 & 10 & 121 & 57 & 570 \\
        \midrule
        \textbf{Total} & 66 & 552 & 58 & 557 & 46 & 441 & 35 & 401 & \textbf{205} & \textbf{1,951} \\
        \bottomrule
    \end{tabular}%
    }
    \label{tab:dataset_stats}
\end{table}
%% Brief statistics
% \begin{table}[tb]
% \caption{Statistics of Sci-ImageMiner Dataset.}
% \centering
% \scriptsize
%     \begin{tabular}{l r r}
%     \toprule
%     \textbf{Split Name} & \textbf{ Papers} & \textbf{ Figures} \\
%     \midrule
%     Train & 126 & 1170 \\
%     Dev & 20 & 201 \\
%     Test & 59 & 580 \\
%     \midrule
%     Total & 205 & 1951 \\
%     \bottomrule
%     \end{tabular}
% \label{tab:dataset_stats}
% \end{table}

% ---- ANNOTATIONS ----
\subsection{Annotations}
\label{sec:dataset_annotations}

% During the dataset curation process, 
We observed that Sci-ImageMiner is a large-scale, inherently complex dataset that requires coordinated annotation across four progressive tasks of scientific comprehension and reasoning. Existing annotation tools proved insufficient to support the domain-specific, end-to-end workflow.
% spanning multiple complementary tasks. 
To address this limitation, we developed a custom web-based, multi-user 
% \href{https://anonymous.4open.science/r/sci-imageminer-annotation-tool-AB3D/}{annotation platform}~\footnote{\url{https://anonymous.4open.science/r/sci-imageminer-annotation-tool-AB3D/}}, 
\href{https://github.com/sciknoworg/sci-imageminer/tree/main/annotation-tool}{annotation platform}~\footnote{\url{https://github.com/sciknoworg/sci-imageminer/tree/main/annotation-tool}}, 
hosted on our institutional infrastructure, that unifies the entire annotation workflow 
% The platform enables 
enabling fine-grained labeling at the subfigure level
% using an incremental alphabetical scheme, with each subfigure 
% localized via 
with bounding-box coordinates 
and a unified JSON schema.
% All annotations are stored in a unified JSON schema that integrates task-specific labels with spatial metadata and is organized within a hierarchical directory structure for efficient management as illustrated in Figure \ref{fig:dataset_organization_hierarchy}.
% % as illustrated below:
% \begin{figure}[htb]
% % \begin{figure}[H]
%   \centering
%   \includegraphics[width=0.7\linewidth]{figures/annotation-workflow-v2.png}
%   \caption{Overview of the Sci-ImageMiner benchmark end-to-end curation workflow.}
%   \label{fig:annotation-workflow}
% \end{figure}
% To further streamline the process as illustrated in Figure~\ref{fig:annotation-workflow}, 
We leveraged \href{https://huggingface.co/Qwen/Qwen2.5-VL-7B-Instruct}{Qwen2.5-VL-7B-Instruct}~\cite{qwen2.5-VL} 
% was leveraged  
for automated pre-annotation for classification, data extraction, and summarization tasks, while the VQA task was intentionally reserved for manual expert annotations to ensure high-quality grounded reasoning.
% in domain expertise.
% Annotation process was
% We recruited 
We recruited 10 domain experts, having postgraduate to postdoctoral qualifications, for the annotation process following a screening phase. Each annotator was assigned 25 papers,
% with a shared 
sharing a subset of 5 papers used to evaluate inter-annotator agreement (IAA). 
% as shown in Tables~\ref{tab:dataset_annotator_stats} and \ref{tab:dataset_annotator_stats}. 
We measured IAA using 
% IAA using 
Fleiss’ Kappa, 
% ~\cite{fleiss1971measuring} 
yielding a score of 0.46 (moderate agreement) on a small 5 paper shared sample. 
% An overview of annotator statistics is presented in Table~\ref{tab:annotated_figures_stats}.
% , indicating moderate agreement, with the relatively low value attributable in part to sample size sensitivity.
Finally, all annotations underwent bi-weekly validations using automated scripts, 
% consistency 
before being consolidated into a structured, distributable dataset. Inconsistencies in annotations were reported to each annotator following the periodic automated validations to align with the annotator guidelines.
% while keeping a fair representation of each annotator across ALD/E and experimental/simulation studies in the dataset.

\begin{table}[htb]
\caption{Top-20 annotated figures in Sci-ImageMiner benchmark dataset (count and percentage).}
\centering
\scriptsize
\setlength{\tabcolsep}{3pt}
\renewcommand{\arraystretch}{1.05}
  \resizebox{\columnwidth}{!}{%
    \begin{tabular}{l c c c c c}
    \toprule
    \textbf{Type} & multiple line chart & molecular structure diagram & line chart & unknown & spectra chart \\
    \textbf{Num} & 716 (15.31\%) & 644 (13.77\%) & 522 (11.16\%) & 374 (8\%) & 343 (7.33\%) \\
    \midrule
    \textbf{Type} & image panel & conceptual diagram & scatter plot & stacked spectra chart & reaction scheme \\
    \textbf{Num} & 324 (6.93\%) & 320 (6.84\%) & 274 (5.86\%) & 266 (5.69\%) & 202 (4.32\%) \\
    \midrule
    \textbf{Type} & multi spectra chart & multi-axis chart & heatmap & apparatus diagram & bar chart \\
    \textbf{Num} & 199 (4.25\%) & 140 (2.99\%) & 73 (1.56\%) & 64 (1.37\%) & 47 (1\%) \\
    \midrule
    \textbf{Type} & contour heatmap & process flow diagram & band diagram & process timing diagram & grouped bar chart \\
    \textbf{Num} & 32 (0.68\%) & 28 (0.6\%) & 22 (0.47\%) & 19 (0.41\%) & 13 (0.28\%) \\
    \bottomrule
    \end{tabular}%
}
\label{tab:annotated_figures_stats}
\end{table}
% \subsubsection{Question Types}
% \label{sec:question_types}

% examples of domain-specific challenging figures
\begin{figure}[!htb]
% \begin{figure}[H]
\centering
% subfigure 1
\begin{subfigure}[t]{0.47\columnwidth}
\centering
\includegraphics[width=0.7\linewidth]{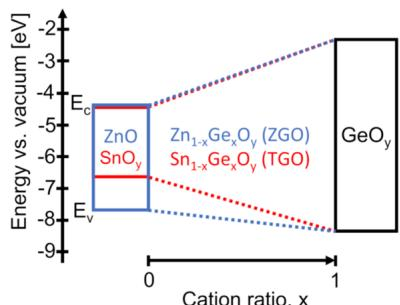}
% \caption{Band Diagram. VQA: structure-property (paragraph). Q: How is the conduction band expected to shift with Ge content? Reasoning: Ideally, the resulting ternary \(Zn_{1-x}Ge_xO_y (ZGO)\) and \(Sn_{1-x}Ge_xO_y (TGO)\) compounds should have Ec positions that shift to higher energies as the Ge content of the compounds increases.}
\caption{Band Diagram. Question-type: Structure-property. Answer-type: Paragraph. Q: How is the conduction band expected to shift with Ge content? Answer: Ideally, the resulting ternary \(Zn_{1-x}Ge_xO_y (ZGO)\) and \(Sn_{1-x}Ge_xO_y (TGO)\) compounds should have Ec positions that shift to higher energies as the Ge content of the compounds increases.}
\end{subfigure}
% subfigure 2
\hfill
\begin{subfigure}[t]{0.47\columnwidth}
\centering
\includegraphics[width=0.7\linewidth]{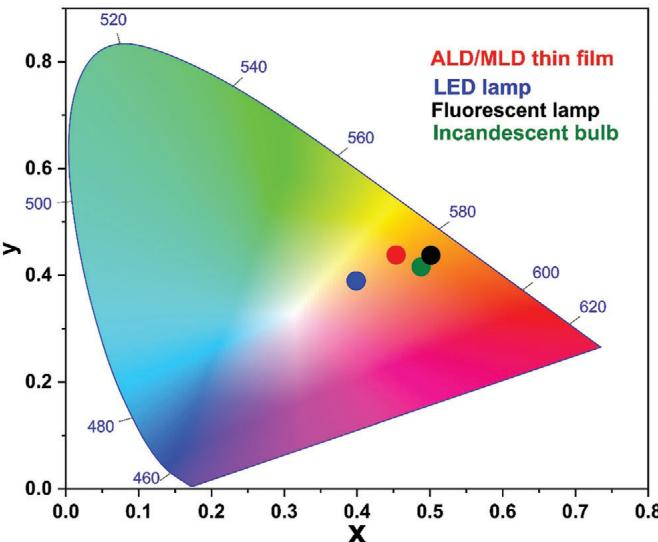}
% \caption{Chromaticity Diagram. VQA: structure-property (factoid). Q: Is the ALD/MLD thin film point closer in color to the incandescent bulb point or to the LED lamp point? Reasoning: It is closer to the incandescent bulb point.}
\caption{Chromaticity Diagram. Question-type: Structure-property. Answer-type: Factoid. Q: Is the ALD/MLD thin film point closer in color to the incandescent bulb point or to the LED lamp point? Answer: It is closer to the incandescent bulb point.}
\end{subfigure}
% \vspace{1em} % Add vertical space between rows
% subfigure 3
\begin{subfigure}[t]{0.47\columnwidth}
\centering
\includegraphics[width=0.7\linewidth]{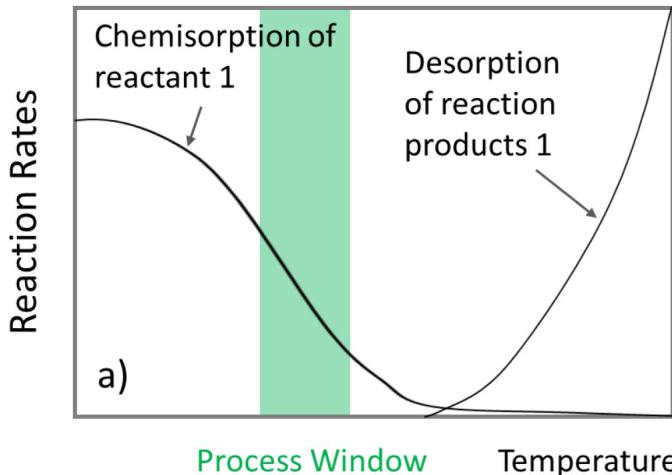}
% \caption{Phase Diagram. VQA: comparative/trend (paragraph). Q: How does the required temperature for the modification step compare to that for the etching step in this process? Reasoning: In thermal isotropic ALE, both steps must occur at the same constant temperature. The process window is the single temperature where modification is complete and etching is spontaneous, making control more challenging than in processes with separate temperature steps.}
\caption{Phase Diagram. Question-type: Comparative/trend. Answer-type: Paragraph. Q: How does the required temperature for the modification step compare to that for the etching step in this process? Answer: In thermal isotropic ALE, both steps must occur at the same constant temperature. The process window is the single temperature where modification is complete and etching is spontaneous, making control more challenging than in processes with separate temperature steps.}
\end{subfigure}
% subfigure 4
\hfill
\begin{subfigure}[t]{0.47\columnwidth}
\centering
\includegraphics[width=0.7\linewidth]{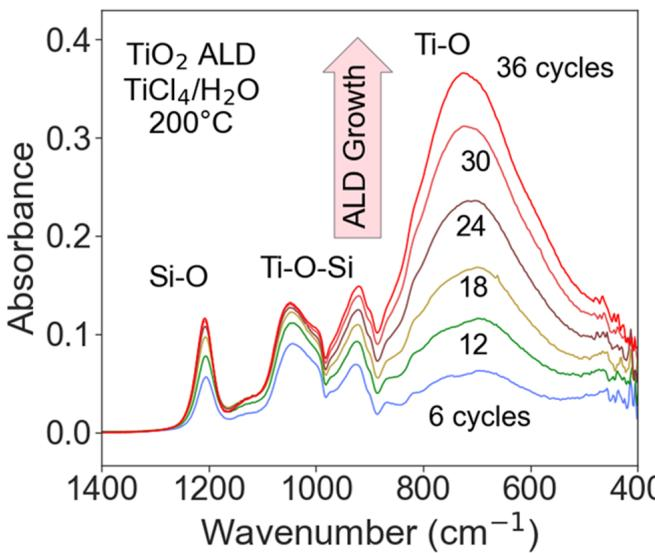}
% \caption{Multi-Spectra Chart. VQA: structure-property (paragraph). Q: How does the Ti-O-Si peak intensity relate to the number of ALD cycles, and what does this signify about film growth? Reasoning: The Ti-O-Si peak grows with increasing cycles, confirming the formation of a chemically bonded \(TiO_2\) film on the \(SiO_2/Si\) substrate. This indicates successful nucleation and steady, layer-by-layer film growth.}
\caption{Multi-Spectra Chart. Question-type: Structure-property. Answer-type: Paragraph. Q: How does the Ti-O-Si peak intensity relate to the number of ALD cycles, and what does this signify about film growth? Answer: The Ti-O-Si peak grows with increasing cycles, confirming the formation of a chemically bonded \(TiO_2\) film on the \(SiO_2/Si\) substrate. This indicates successful nucleation and steady, layer-by-layer film growth.}
\end{subfigure}
\caption{Examples of some \textit{challenging} scientific figures with domain-grounded VQA in Sci-ImageMiner.}
\label{fig:domain_specific_figure_examples}
\end{figure}

% Specifically 
In particular, for the VQA, the annotators assigned four question-answer pairs per figure, labeled at the subfigure level as shown in~\autoref{fig:domain_specific_figure_examples}. 
\textbf{Question Types} are guided by Bloom’s~\cite{revisedbloomtaxonomy} revised taxonomy 
% of cognitive processes, 
to progressively assess deeper levels of understanding.
% , ranging from basic pattern recognition to advanced reasoning involving mechanistic interpretation and application-level inference as shown in the Table~\ref{tab:vqa_question_types}
\textbf{Process--Oriented} 
% focuses on understanding 
reaction mechanisms, experimental workflows.
% , requiring application-level reasoning over sequential steps.
% and the ability to infer causal relationships within ALD/E processes.
\textbf{Comparative/Trend} 
% evaluates the interpretation of 
quantitative relationships, correlations.
% , inflection points and performance variations across conditions.
\textbf{Structure--Property} 
% assesses the ability to 
connects chemical composition,  
% precursor 
structures to resulting material properties.
% , with analytical and application-level reasoning.
\textbf{Application/Performance} 
% translates 
experimental observations to device-level or application-oriented insights.
\textbf{Answer Types} follow four standardized formats, 
% allowing evaluation of both concise retrieval and explanatory reasoning for each question type.
\textbf{Yes/No}: exclusive disjunction.
% i.e. either "Yes" or "No".
\textbf{Factoid}: brief factual answer.
% (e.g., "O$_2$ plasma").
\textbf{List}: comma-separated order-insensitive list of answers. \textbf{Paragraph}: explanatory answer providing detailed insights.
\section{Competition}
\label{sec:competition}

% In this section, we present the Sci-ImageMiner ICDAR competition. In Section~\ref{sec:competition_timeline_schedule}, we discuss the competition timeline schedule. Section~\ref{sec:competition_organization} describes how the competition was organized in detail. Finally, Section~\ref{sec:competition_tasks} discuss all the four competition task descriptions and evaluation protocols.

\subsection{Timeline Schedule}
\label{sec:competition_timeline_schedule}
% The timeline of the Sci-ImageMiner competition is summarized as:
%(all times are in AoE):
% \begin{itemize}
%     \item Competition website launch: 28th November 2025
%     \item Call for Participation: 12th December 2025
%     \item Registrations Open: 6th January 2026
%     \item Development Phase Begin: 9th January 2026
%     \item Training dataset release (Batch 1): 14th January 2026
%     \item Training dataset release (Batch 2): 5th January 2026
%     \item Full Train/Dev and Blind Test data release: 6th March 2026
%     \item Development Phase End: 15th March 2026
%     \item Evaluation Phase Begin: 16th March 2026
%     \item Evaluation Phase End: 8th April 2026
% \end{itemize}
% The timeline of the ICDAR competition was structured to support progressive participation and systematic evaluation.
The competition website was launched on 28 November 2025,
% followed by the official call for participation on 12 December 2025 and the opening of registrations on 6 January 2026.
the development phase commenced on 9 January 2026, accompanied by progressive releases of the training dataset in batches,
% with Batch 1 on 14 January 2026 and Batch 2 made available on 5 February 2026.
subsequently, the complete train/dev and blind test datasets were released on 6 March 2026.
% , enabling participants to finalize their systems.
The 
% development phase concluded on 15 March 2026, after which the
evaluation phase began on 16 March 2026 and continued until 8 April 2026.
% , marking the end of the competition cycle.

% ---- ORGANIZATION ----
\subsection{Competition Organization}
\label{sec:competition_organization}

A dedicated competition website~\footnote{\url{https://sites.google.com/view/sci-imageminer/home}}
was established,
% to provide 
providing comprehensive information
% , detailed task descriptions, evaluation methodologies, submission guidelines, 
% and supporting resources. The Sci-ImageMiner dataset and evaluation scripts were publicly released and are hosted on the official GitHub repository~\footnote{\url{https://github.com/sciknoworg/ALD-E-ImageMiner/tree/main/icdar2026-competition-data}}.
and supporting resources.
To ensure effective communication with participants, 
% we established 
a dedicated 
% \href{https://groups.google.com/g/sci-imageminer}{Google Group} 
Google Group~\footnote{\url{https://groups.google.com/g/sci-imageminer}} 
was established 
% through which
for periodic announcements and Q\&A.
% The Sci-ImageMiner dataset and evaluation scripts were publicly released and are hosted on the official GitHub repository~\footnote{\url{https://github.com/sciknoworg/sci-imageminer/}}.
% The Sci-ImageMiner dataset and evaluation scripts are hosted on the official GitHub repository~\footnote{\url{https://github.com/sciknoworg/sci-imageminer/}}.
% , ensuring accessibility and reproducibility.
% ~\footnote{\url{https://groups.google.com/g/sci-imageminer}}
% Task 1~\footnote{\url{https://www.codabench.org/competitions/12901/}}
% Task 2~\footnote{\url{https://www.codabench.org/competitions/12902/}}
% Task 3~\footnote{\url{https://www.codabench.org/competitions/12909/}}
% Task 4~\footnote{\url{https://www.codabench.org/competitions/12908/}}
% To facilitate efficient submission handling and evaluation, 
% we leveraged 
The \href{https://www.codabench.org}{CodaBench} platform was utilized by organizing separate competition tracks for each task, to allow 
% \href{https://www.codabench.org/competitions/12901/}{Task 1}, \href{https://www.codabench.org/competitions/12902/}{Task 2}, \href{https://www.codabench.org/competitions/12909/}{Task 3}, and \href{https://www.codabench.org/competitions/12908/}{Task 4}.
% This 
% modular setup enabled 
scalable and well-structured management of submissions and evaluations without manual intervention using our evaluation scripts.
% Evaluation scripts were developed to support automated evaluation on CodaBench.
% , we developed a custom Docker container that provides a standardized compute environment with the necessary dependencies for executing evaluation scripts and generating leaderboard scores.

% were disseminated
% effectively communicated. 
The competition was conducted in two phases: (i) \textbf{development phase} 
% , which 
allowed participants to design, train, and validate their approaches, while (ii) \textbf{evaluation phase} 
% , during which 
allowed assessments of final submissions 
% were assessed and ranked
and rankings on the official leaderboards. 
% To maintain fairness, duplicate submissions from the same team were restricted, and only the best-performing submission per task was considered for final ranking.
Following the 
% completion of the 
competition, 
% The 
% top five teams from each task leaderboard 
the winning teams were invited to submit system papers to the 
% Sci-ImageMiner 
% \href{https://doi.org/10.52825/ocp.v10i}{proceedings}~\footnote{\url{https://doi.org/10.52825/ocp.v10i}} 
proceedings~\footnote{\url{https://doi.org/10.52825/ocp.v10i}} via 
% \href{https://www.tib-op.org/ojs/}{TIB Open Conference Publishing} free of charge.
TIB Open Conference Publishing~\footnote{\url{https://www.tib-op.org/ojs/}} free of charge.

% ---- TASKS ----
\subsection{Tasks}
\label{sec:competition_tasks}

% The Sci-ImageMiner dataset is curated to support four key tasks that collectively evaluate end-to-end comprehensive scientific figure understanding and reasoning capabilities.
The competition is organized as such to 
% four end-to-end key tasks that 
collectively evaluate end-to-end comprehensive scientific figure understanding and reasoning capabilities.

\subsubsection{Task 1: Classification}
\label{sec:task_cls}

\hfill \break \textit{Task Description:} 
% The objective of this task is 
% To classify each figure into one of the 49 predefined categories defined in the \href{https://github.com/sciknoworg/ALD-E-ImageMiner/blob/main/figure_taxonomy.png}{taxonomy}. \\
To classify each figure into one of the 49 predefined categories defined in the \href{https://github.com/sciknoworg/sci-imageminer/blob/main/figure_taxonomy.png}{taxonomy}. \\
% which includes 28 qualitative plot types. \\
\textit{Evaluation Protocol:} Performance is evaluated using Accuracy, Precision, Recall, and F1-score. The F1-score serves as the primary metric for ranking teams on the leaderboard.

\subsubsection{Task 2: Data Extraction}
\label{sec:task_dte}

\hfill \break \textit{Task Description:} 
% This task requires participants 
To reconstruct the underlying data represented in a chart by extracting it into a structured Markdown table format. \\
\textit{Evaluation Protocol:} The evaluation is based on Relative Mapping Similarity (RMS) and Tree Edit Distance-based Similarity (TEDS). The final ranking score is computed as a weighted combination of these metrics, defined as 
$\frac{1}{2}(\text{RMS}) + \frac{1}{2}(\text{TEDS})$.
% $\frac{1}{2} \times \text{RMS}$
% ($0.5 \times \text{RMS}) + (0.5 \times \text{TEDS})$.

\subsubsection{Task 3: Summarization}
\label{sec:task_summ}

\hfill \break \textit{Task Description:} 
% The goal of this task is 
To generate concise and factually accurate summaries that capture the key 
% trends, relationships and 
insights. \\
% conveyed by the figure.
% Participants may optionally leverage the associated figure caption and relevant textual context to enhance the quality and factual grounding of the summaries. \\
\textit{Evaluation Protocol:} Evaluation is conducted using ROUGE-1/2/L
% ~\cite{lin2004rouge}
, and BERTScore-F1
% ~\cite{zhang_bertscore_2020}
. The final score is computed as a weighted combination of these metrics, defined as 
$\frac{1}{2}(\frac{R1 + R2 + RL}{3}) + \frac{1}{2}(\text{BERTScore-F1})$. 
% $(0.5 \times \text{Average(ROUGE-1/2/L)}) + (0.5 \times \text{BERTScore-F1})$, 
% and is used to rank teams on the leaderboard.
%% NO subsubsections, use \noindent{\textbf{blah blah}}
\subsubsection{Task 4: Visual Question Answering}
\label{sec:task_vqa}

\hfill \break \textit{Task Description:} 
% This task evaluates 
Domain-specific reasoning over expert-annotated question-answer pairs. \\
% , spanning four categories: Comparative/Trend analysis, Structure-Property reasoning, Process-Oriented reasoning, and Application/Performance assessment. Each question is associated with one of four answer types: Yes/No, Factoid, List, or Paragraph. Similar to the summarization task.
% Participants may utilize associated figure captions and contextual text to improve answer quality and factual consistency. \\
\textit{Evaluation Protocol:} Evaluation is conducted individually for each answer type. Paragraph: assessed using ROUGE-1/2/L and BERTScore-F1. Factoid: evaluated using the ROUGE-1/2/L and Exact Match. List: evaluated using set-based F1 scores with case-insensitive matching. Yes/No: evaluated using Accuracy and F1-score. The final ranking score is computed as a weighted score, 
% combination of all answer types, 
defined as 
$\frac{1}{4}(\text{Paragraph}) + \frac{1}{4}(\text{Factoid}) + \frac{1}{4}(\text{List}) + \frac{1}{4}(\text{Yes/No})$.
% $(0.25 \times \text{Paragraph}) + (0.25 \times \text{Factoid}) + (0.25 \times \text{List}) + (0.25 \times \text{Yes/No})$.

% ---- RESULTS ----
\section{Results and Discussion}
\label{sec:results_and_discussions}

\subsection{Overview}
\label{sec:results_and_discussions_overview}

The ICDAR Sci-ImageMiner attracted substantial engagement from around the globe, with a total of 68 active participants contributing 1,263 public and private submissions
% across all competition leaderboards
as shown
% spanning all four tasks 
% Detailed statistics regarding participation and submission distributions are summarized
in Table~\ref{tab:competition_stats}.
% Following the conclusion of the submission phase, the 
% Top-5 teams from each 
% task-specific 
% leaderboard~\ref{tab:results_winning_teams} 
% were identified based on the designated primary evaluation metric, or a composite weighted score in cases involving multiple metrics, 
% ensuring a fair and consistent ranking protocol.
% The complete leaderboards for all tasks are publicly accessible via the respective CodaBench competition pages: \href{https://www.codabench.org/competitions/12901/}{Task 1}, \href{https://www.codabench.org/competitions/12902/}{Task 2}, \href{https://www.codabench.org/competitions/12909/}{Task 3}, and \href{https://www.codabench.org/competitions/12908/}{Task 4}.
The \href{https://sites.google.com/view/sci-imageminer/team-results-leaderboard}{full leaderboards} are available on our competition website.

%% Show stats for dev and eval phases in the same table
% \begin{table}[htb]
\begin{table}[htb]
    \centering
    \caption{The statistics of the competition participants with all private and public submissions}
    \scriptsize
    % \setlength{\tabcolsep}{3pt}
    % \renewcommand{\arraystretch}{1.05}
    % \resizebox{\columnwidth}{!}{%
    \begin{tabular}{l c c}
    \toprule
    \textbf{Task} & \textbf{ Participants} & \textbf{Submissions} \\
    \midrule
    Classification &  81 &  501 \\
    Data Extraction & 53 &  335 \\
    Summarization & 43 &  212 \\
    Visual Question-Answering & 47 &  215 \\
    \midrule
    TOTAL &  ~ & 1,263 \\
    \bottomrule
    \end{tabular}%
    % }
\label{tab:competition_stats}
\end{table}

\subsection{Baseline Results}
\label{baseline_results}

We established baselines 
% systems for all four Sci-ImageMiner tasks 
by evaluating a range of 
% general-purpose 
LVLMs, including 
% \href{}{}, 
\href{https://huggingface.co/google/gemma-4-E4B-it}{Gemma 4 E4B 8b}, 
\href{https://huggingface.co/Qwen/Qwen3-VL-8B-Instruct}{Qwen3-VL-8B-Instruct}~\cite{qwen3technicalreport}, 
% ~\cite{qwen3technicalreport}
\href{https://huggingface.co/zai-org/GLM-4.6V-Flash}{GLM-4.6V-Flash}~\cite{vteam2025glm45vglm41vthinkingversatilemultimodal}, 
\href{https://huggingface.co/OpenGVLab/InternVL3_5-8B}{Intern VL 3.5 8b}~\cite{wang2025internvl3_5}, 
% % and \href{https://developers.openai.com/api/docs/models/gpt-5-mini}{GPT-5.1 mini}, 
% alongside chart-specialized VLMs such as \href{https://huggingface.co/google/deplot}{DePlot}
% % ~\cite{liu2022deplot}
% and \href{https://huggingface.co/ahmed-masry/chartgemma}{ChartGemma} 
% ~\cite{masry2024chartgemma}.
% All models were evaluated under a unified protocol, same as the ICDAR competition, with carefully designed
% We 
% , and 
while leveraging the 
% \href{https://github.com/sciknoworg/sci-imageminer/tree/main/baseline/prompts}{prompts} 
prompts~\footnote{\url{https://github.com/sciknoworg/sci-imageminer/tree/main/baseline/prompts}} 
to ensure consistency across tasks and baselines.
Experimental results indicate that 
% % both general-purpose and chart-specialized 
% % baselines perform sub-optimally on this dataset. Notably, 
% chart-specialized VLMs
% % despite strong performance on standard benchmarks, 
% fail to generalize to complex scientific figures for the data extraction task, while LVLMs underperformed across all
% % end-to-end reasoning 
% tasks. 
no single baseline performed well across all tasks, 
% These findings 
highlighting the intrinsic difficulty of scientific figure comprehension and the need for more advanced multimodal reasoning capabilities tailored to domain-specific complexities.
% (i) Classification was assessed using Accuracy and F1-score; (ii) Data Extraction was evaluated using RMS~\cite{liu2022deplot} for numerical accuracy and TEDS~\cite{zhong2019image} for structural similarity; (iii) Summarization was measured using ROUGE-1/2/L~\cite{lin2004rouge} for lexical overlap and BERTScore-F1~\cite{zhang_bertscore_2020} for semantic similarity; and (iv) Visual Question Answering (VQA) was evaluated in a task-specific manner, including Yes/No (Accuracy/F1), Factoid (ROUGE-1/2/L and Exact Match), Paragraph (ROUGE-1/2/L and BERTScore), and List (set-F1).

\subsection{Competition Results}
\label{competition_results}

In this section, we summarize the methodologies proposed by 
% participating teams.
% emphasizing the diversity of approaches across tasks. 
% The discussion is restricted to 
the top-5 leaderboard teams, based on the availability of corresponding method reports. An overview of all the top-performing team approaches is shown in Table~\ref{tab:results_tldr}.

\subsubsection{Task 1: Classification Results}
\label{sec:results_cls}

The approaches used by the top-5 submissions are presented below, and the Table~\ref{tab:leaderboard_classification} shows the task leaderboard.

\begin{enumerate}
    \item \textbf{Ricoh\_SRCB} \\
    The team employs a hierarchical fine-tuning strategy, initially training a coarse-grained 6-class model followed by a refined 49-class model, complemented by auxiliary binary and 4-class classifiers to address particularly challenging categories. Final predictions are obtained by fusing results from these models. At inference time, performance is further enhanced by incorporating both global and locally cropped image inputs, enabling the model to capture complementary contextual and fine-grained visual features.
    % This multi-stage and multi-view design achieves strong performance, with 0.79 accuracy and 0.81 F1-score.
    % \item \textbf{pana19} \\
    % % TODO: Add team specific approaches for the task
    % The team did not submit details of their approach yet.

    \item \textbf{IIT\_PATNA\_CV\_1} \\
    This team reformulates figure panel classification as a generative task by fine-tuning the vision-language model \href{https://huggingface.co/Qwen/Qwen2.5-VL-7B-Instruct}{Qwen2.5-VL-7B-Instruct}~\cite{qwen2.5-VL} using Low-Rank Adaptation (LoRA). The model is prompted with both the input image and a taxonomy-aware textual description, enabling direct generation of class labels without an explicit classification head. Robustness is improved through class-aware oversampling and parameter-efficient training, while inference leverages test-time augmentation and majority voting to enhance prediction stability. A post-hoc label mapping ensures alignment with the predefined taxonomy.
    % This unified generative framework yields competitive performance, achieving 0.76 accuracy and 0.77 F1-score.

    \item \textbf{DocMiner} \\
    The team's proposed approach emphasizes enriched contextual modeling by augmenting inputs with category-specific sample documents, image captions, and source text to improve semantic understanding. An iterative prompt optimization process, guided by development set feedback, refines category descriptions with particular focus on resolving ambiguities between visually similar classes. Additionally, a multi-agent inference framework is introduced, in which independent predictions are reconciled through consensus or re-evaluation mechanisms to mitigate individual-model bias.
    % This combination of contextual enrichment and collaborative decision-making results in robust classification performance, with 0.75 accuracy and 0.76 F1-score.
    
    \item \textbf{VLMinators} \\
    This team's approach adopts a systematic experimental framework, benchmarking multiple vision-language models and CNN architectures before identifying QLoRA-based~\cite{dettmers2023qlora} fine-tuning of models such as \href{https://huggingface.co/Qwen/Qwen2.5-VL-7B-Instruct}{Qwen2.5-VL-7B-Instruct}~\cite{qwen2.5-VL} and LLaMA-based vision models as the most effective configuration. A key innovation is the use of selective label hints-targeted textual disambiguation cues embedded in prompts, to improve differentiation between confusable classes. Additional gains are achieved through post-processing pipelines for label normalization and limited manual correction of uncertain predictions. Notably, ensemble methods across heterogeneous architectures proved ineffective due to high inter-model disagreement.
    % The final system achieves 0.74 accuracy and 0.75 F1-score.
\end{enumerate}
% TODO: Insert table of Leaderboard scores for the task
% \begin{table}[tb]
% \centering
% \scriptsize
%     \begin{tabular}{l l c c c c}
%     \toprule
%     \textbf{\#} & \textbf{Team} & \textbf{Accuracy} & \textbf{Precision} & \textbf{Recall} & \textbf{F1-score} \\
%     \midrule
%     % 1 & Ricoh\_SRCB & 0.79 & 0.82 & 0.80 & 0.81 \\
%     1 & RSRCB & 0.79 & 0.82 & 0.80 & 0.81 \\
%     2 & pana19 & 0.76 & 0.78 & 0.77 & 0.77 \\
%     % 3 & IIT\_PATNA\_CV\_1 & 0.76 & 0.79 & 0.75 & 0.77 \\
%     3 & IIT PATNA & 0.76 & 0.79 & 0.75 & 0.77 \\
%     % 4 & Zoloz-RealDoc & 0.75 & 0.76 & 0.76 & 0.76 \\
%     4 & SMU & 0.75 & 0.76 & 0.76 & 0.76 \\
%     5 & CMU & 0.74 & 0.75 & 0.75 & 0.75 \\
%     \midrule
%     6 & GPT 5.1 Mini (baseline) & 0.27 & ~ & ~ 0.37 \\
%     \bottomrule
%     \end{tabular}
% \caption{Task 1 Classification scores for the top-5 teams}
% \label{tab:leaderboard_classification}
% \end{table}
\begin{table}[htb]
    \centering
    \caption{Task 1 Classification scores for the best baseline and top-5 teams}
    \scriptsize
    \begin{tabular}{l l c c c c}
    \toprule
    \textbf{\#} & \textbf{Team} & \textbf{Accuracy} & \textbf{F1-score} \\
    \midrule
    1 & Ricoh\_SRCB & 0.79 & 0.81 \\
    2 & BOE AI+ & 0.76 & 0.77 \\
    3 & IIT\_PATNA\_CV\_1 & 0.76 & 0.77 \\
    4 & DocMiner & 0.75 & 0.76 \\
    5 & VLMinators & 0.74 & 0.75 \\
    \midrule
    % 6 & GPT 5.1 Mini (baseline) & 0.27 & 0.37 \\
    6 & \href{https://huggingface.co/google/gemma-4-E4B-it}{Gemma 4 E4B 8b} (baseline) & 0.67 & 0.68 \\
    \bottomrule
    \end{tabular}
\label{tab:leaderboard_classification}
\end{table}

\subsubsection{Task 2: Data Extraction Results}
\label{sec:results_dte}
The approaches used by the top-5 submissions are presented below and 
 Table~\ref{tab:leaderboard_data_extraction} shows the task leaderboard.

\begin{enumerate}
    \item \textbf{TeleOCR-VL} \\
    The team proposes a comprehensive framework for chart-to-structure extraction that integrates structure-aware training, large-scale synthetic data generation, and ensemble-based inference. By explicitly decoupling structural learning from content recognition via separate training on structure-only and full-content samples, the method improves both layout understanding and semantic extraction. The use of synthetic data mitigates annotation scarcity and enhances generalization, while an iterative pseudo-label refinement loop further reduces noise. At inference time, a heterogeneous ensemble balances structural fidelity and numerical accuracy, resulting in consistent and significant performance improvements.
    % , underscoring the importance of both data-centric design and model aggregation strategies.

    % \item \textbf{tirtha-v}
    \item \textbf{VLMinators} \\
    The team adopts a parameter-efficient fine-tuning strategy on \href{https://huggingface.co/Qwen/Qwen2.5-VL-7B-Instruct}{Qwen2.5-VL-7B-Instruct}~\cite{qwen2.5-VL} using QLoRA~\cite{dettmers2023qlora}, focusing on structured table generation from chart images. The approach relies heavily on prompt engineering, enforcing explicit formatting constraints and leveraging semantic cues such as axis labels and legends for column construction. A key contribution is the use of context injection, where chart-type metadata is prepended during inference to guide structural predictions. This enables improved alignment between chart semantics and table schema, effectively reducing structural ambiguities with minimal architectural modification.

    \item \textbf{Ricoh\_SRCB} \\
    The team's approach emphasizes input representation and data balancing by incorporating multi-image prompts that combine full charts with their corresponding cropped subgraphs. Such a design enhances the model’s ability to capture both global and local visual patterns. Additionally, category-wise data augmentation is employed to address class imbalance, ensuring more uniform training coverage across chart types and improving generalization.
    
    % \item \textbf{Vsioros (NCSR Demokritos \& National and Kapodistrian University of Athens)} \\
    % \item \textbf{NCSR Demokritos \& National and Kapodistrian University of Athens (UOA)} \\
    \item \textbf{Vassilis Sioros} \\
    The team utilizes \href{https://huggingface.co/Qwen/Qwen3.5-9B}{Qwen3.5-9B} with advanced architectural components, including early vision-language fusion, Gated Delta Networks, and sparse Mixture-of-Experts. It incorporates enriched contextual inputs by combining cropped subfigures with the surrounding text from the source document. The model generates dense string representations of tables, which are then parsed into structured Markdown. 
    % This design highlights a tightly integrated vision-language pipeline that leverages both local visual cues and broader document semantics for improved extraction fidelity.

    % \item \textbf{Zoloz-RealDoc (Singapore Management University)} \\
    \item \textbf{DocMiner} \\
    The team explicitly integrated auxiliary textual information, such as captions and surrounding document text, into the input to enhance semantic grounding. Furthermore, it introduces a multi-agent architecture in which specialized agents are assigned to distinct question types, with dynamic routing enabling task-specific processing.
    % This modular design improves precision and robustness by decomposing the problem into specialized sub-tasks and leveraging targeted reasoning strategies.
\end{enumerate}

% TODO: Insert table of Leaderboard scores for the task
%% insert the result from the best baseline in the following table as a new row (labeled as baseline in paranthesis),
\begin{table}[htb]
\centering
\scriptsize
    \begin{tabular}{l l c c c}
    \toprule
    \textbf{\#} & \textbf{Team} & \textbf{RMS} & \textbf{TEDS} & \textbf{Weighted} \\
    \midrule
    1 & TeleOCR-VL & 17.23 & 66.39 & 41.81 \\
    % 2 & tirtha-v & 17.29 & 64.31 & 40.80 \\
    2 & VLMinators & 17.29 & 64.31 & 40.80 \\
    3 & Ricoh\_SRCB & 16.23 & 61.12 & 38.67 \\ 
    % 4 & vsioros & 14.94 & 55.20 & 35.07 \\
    % 4 & NDNKU & 14.94 & 55.20 & 35.07 \\
    4 & Vassilis Sioros & 14.94 & 55.20 & 35.07 \\
    % 5 & ZoloZ-RealDoc & 12.67 & 53.72 & 33.19 \\
    5 & DocMiner & 12.67 & 53.72 & 33.19 \\
    \midrule
    % 6 & GPT 5.1 Mini (baseline) & 3.58 & 19.52 & 11.55 \\
    6 & Qwen 3 VL 8b~\cite{qwen3technicalreport} (baseline) & 14.08 & 57.86 & 35.97 \\
    \bottomrule
    \end{tabular}
\caption{Task 2 Data Extraction scores for the best baseline and top-5 teams}
\label{tab:leaderboard_data_extraction}
\end{table}

%% Team methods TLDR;
% \begin{table}[!ht]
\begin{table}[H]
% \begin{table}[htb]
    \centering
    \caption{Overview of top teams, methods, LVLMs used, and rankings from across all competition tasks leaderboards. The \href{https://sites.google.com/view/sci-imageminer/team-results-leaderboard}{full leaderboards} are released on our competition website.}
    \scriptsize
    \setlength{\tabcolsep}{3pt}
    \renewcommand{\arraystretch}{1.05}
    \resizebox{\columnwidth}{!}{%
    \begin{tabular}{l p{5cm} p{2cm} p{1.8cm}}
    \toprule

    \textbf{Team} & \textbf{Method TLDR; } & \textbf{LVLMs Used} & \textbf{Rankings} \\
    \midrule
    DeepVitminC & \textbf{Task3:} RAG-based context extraction with alignment-driven multimodal fine-tuning for consistent VQA and summarization.\newline
    \textbf{Task4:} Rule-based context extraction with RAG-augmented prompting and alignment-focused VLM fine-tuning for semantically grounded outputs. & \textbf{Task3,4:} \href{https://huggingface.co/Qwen/Qwen3.5-4B}{Qwen3.5-4B}; \href{https://huggingface.co/Qwen/Qwen3.5-35B}{Qwen3.5-35B}
    % ~\cite{qwen3.5} 
    & \textbf{Task3:} 1st;\newline \textbf{Task4:} 1st \\ 
    DocMiner & \textbf{Task1:} Context-enriched prompting with iterative refinement and multi-agent consensus for robust classification.\newline
    \textbf{Task2:} Context-augmented multi-agent VLM framework with task-specific routing for robust semantic extraction.\newline
    \textbf{Task4:} Iterative multi-agent VQA framework with task specialization and dev-driven prompt refinement for improved robustness. & \textbf{Task1,2,4:} \href{https://huggingface.co/Qwen/Qwen3.5-397B-A17B}{Qwen3.5-plus}
    % ~\cite{qwen3.5}
    ; \href{https://deepmind.google/models/model-cards/gemini-3-1-pro/}{Gemini 3.1 Pro} & Task1: 4th;\newline \textbf{Task2:} 5th;\newline \textbf{Task4:} 3rd \\ 
    IIT\_PATNA\_CV\_1 & \textbf{Task1:} Generative VLM-based classification with LoRA, taxonomy-aware prompting, class-balanced training, and TTA-based ensembling. & \textbf{Task1:} \href{https://huggingface.co/Qwen/Qwen2.5-VL-7B-Instruct}{Qwen2.5-VL-7B-Instruct} & \textbf{Task1:} 3rd \\ 
    Ricoh\_SRCB & \textbf{Task1:} Multi-stage fine-tuning with hierarchical classifiers and dual-image input augmentation for improved classification accuracy.\newline
    \textbf{Task2:} Multi-image prompting with category-balanced augmentation for enhanced chart representation and generalization.\newline
    \textbf{Task3:} LoRA + DPO-based~\cite{rafailov_direct_2023} VLM fine-tuning with hard negative sampling and structured prompting for improved alignment.\newline \textbf{Task4:} Two-stage SFT + GRPO alignment with limited gains due to data and initialization constraints in preference optimization. & \textbf{Task1,2,3:} \href{https://huggingface.co/Qwen/Qwen3.5-9B}{Qwen3.5-9B}
    % ~\cite{qwen3.5}
    ;\newline  \textbf{Task4:} \href{https://huggingface.co/Qwen/Qwen2.5-VL-7B-Instruct}{Qwen2.5-VL-7B-Instruct} & \textbf{Task1:} 1st;\newline \textbf{Task2:} 3rd;\newline \textbf{Task3:} 2nd;\newline Task4: 2nd \\ 
    TeleOCR-VL & \textbf{Task:2:} Structure-aware chart-to-structure extraction with synthetic data generation, pseudo-label refinement, and ensemble-based inference.\newline 
    \textbf{Task3:} Vision-context dual-stream summarization with chart priors and consensus-based re-ranking for factual consistency. & \textbf{Task2,3:} TeleOCR-VL & \textbf{Task2:} 1st;\newline \textbf{Task3:} 3rd \\ 
    Vassilis Sioros & \textbf{Task2:} Advanced VLM architecture with early fusion, MoE, and context-enriched inputs for end-to-end chart-to-table extraction.\newline 
    \textbf{Task3:} Scalable early-fusion VLM with MoE, sliding-window context retrieval, and data-normalized fine-tuning for robust summarization.\newline \textbf{Task4:} Unified multimodal VLM fine-tuning with hierarchical evidence generation and format-aware preprocessing for enhanced VQA reasoning. & \textbf{Task2,3,4:} \href{https://huggingface.co/Qwen/Qwen3.5-9B}{Qwen3.5-9B}
    % ~\cite{qwen3.5}
    & Task2: 4th;\newline \textbf{Task3:} 5th;\newline \textbf{Task4:} 4th \\ 
    VLMinators & \textbf{Task1:} QLoRA-tuned~\cite{dettmers2023qlora} VLMs with selective label hints and post-processing outperform baselines, while ensembling is hindered by inter-model disagreement.\newline 
    \textbf{Task2:} QLoRA-based~\cite{dettmers2023qlora} VLM fine-tuning with prompt engineering and context injection for structured table generation.\newline 
    \textbf{Task4:} QLoRA-based~\cite{dettmers2023qlora} VLM fine-tuning with answer-type-aware prompting and sequential context chaining for cross-task enhanced VQA. & \textbf{Task1,2,4:} \href{https://huggingface.co/Qwen/Qwen2.5-VL-7B-Instruct}{Qwen2.5-VL-7B-Instruct}
    % ~\cite{qwen2.5-VL}
    ; \textbf{Task1:} \href{https://huggingface.co/meta-llama/Llama-3.2-11B-Vision-Instruct}{Llama-3.2-11B-Vision-Instruct} & \textbf{Task1:} 5th;\newline \textbf{Task2:} 2nd;\newline \textbf{Task4:} 5th \\
    \bottomrule
    \end{tabular}%
    }
    \label{tab:results_tldr}
\end{table}

\subsubsection{Task 3: Summarization Results}
\label{sec:results_summ}
The approaches used by the top-5 submissions are presented below and 
Table~\ref{tab:leaderboard_summarization} shows the task leaderboard.

\begin{enumerate}
    \item \textbf{DeepVitminC} \\
    The team introduces a modular framework composed of a Context Extraction Module and a Consistent Alignment Module to support summarization and visual question answering. Contextual information is derived from the source document, primarily via figure captions using retrieval-augmented generation and injected into the model input. The alignment module then jointly processes visual and contextual signals to produce outputs that conform to task-specific response formats, learned through supervised fine-tuning.
    % Overall, the method emphasizes structured context integration and end-to-end alignment for improving semantic coherence and output consistency.
    
    \item \textbf{Ricoh\_SRCB} \\
    This team's approach builds on \href{https://huggingface.co/Qwen/Qwen2.5-VL-7B-Instruct}{Qwen2.5-VL-7B-Instruct}~\cite{qwen2.5-VL}, using LoRA-based fine-tuning followed by Direct Preference Optimization (DPO)~\cite{rafailov_direct_2023} to enhance alignment with ground-truth answers. A key contribution lies in the construction of hard negative samples through stochastic corruption of correct answers, which proved more effective than weaker model-generated outputs for preference learning. Additionally, prompt design incorporates captions, explicit subgraph enumeration, and strict formatting constraints, collectively improving structured reasoning.
    % The approach highlights the importance of preference optimization and carefully designed negative sampling in refining model behavior.
    
    \item \textbf{TeleOCR-VL} \\
    The team's approach targets chart summarization by integrating visual inputs with rich contextual information, including captions and surrounding manuscript text, within a dual-stream "vision + context" paradigm. Chart-type priors are further embedded to guide domain-specific language generation. During inference, multiple candidate summaries are generated and re-ranked based on semantic similarity, consensus scoring, and cross-modal validation against structured data, with inconsistent outputs filtered through conflict resolution.
    % Empirical gains demonstrate that contextual grounding and consensus-based selection significantly enhance factual accuracy and coherence.

    % \item \textbf{pana19} \\
    % % TODO: Add team specific approaches for the task
    % The team did not submit details of their approach yet.

    % \item \textbf{Vsioros (NCSR Demokritos \& National and Kapodistrian University of Athens)} \\
    % \item \textbf{NCSR Demokritos \& National and Kapodistrian University of Athens (NDNKU)} \\
    \item \textbf{Vassilis Sioros} \\
    This team's approach leverages \href{https://huggingface.co/Qwen/Qwen3.5-9B}{Qwen3.5-9B} with an early-fusion multimodal architecture augmented by Gated Delta Networks and sparse Mixture-of-Experts. It combines cropped subfigure inputs with localized textual context retrieved via a sliding window over the document. Training data is carefully normalized and filtered under strict token-budget constraints to ensure consistent format and efficiency.
    % The approach underscores the role of architectural scalability and disciplined data preprocessing in achieving robust multimodal summarization performance.

\end{enumerate}

% TODO: Insert table of Leaderboard scores for the task
\begin{table}[htb]
\centering
\scriptsize
    \begin{tabular}{l l c c c c c}
    \toprule
    \textbf{\#} & \textbf{Team} & \textbf{R1} & \textbf{R2} & \textbf{RL} & \textbf{BERT-F1} & \textbf{Weighted} \\
    \midrule
    1 & DeepVitminC & 0.37 & 0.17 & 0.30 & 0.84 & 0.56 \\
    2 & Ricoh\_SRCB & 0.37 & 0.16 & 0.29 & 0.83 & 0.55 \\
    3 & TeleOCR-VL & 0.35 & 0.15 & 0.28 & 0.82 & 0.54 \\
    4 & BOE AI+ & 0.35 & 0.15 & 0.28 & 0.82 & 0.54 \\
    % 5 & NDNKU & 0.34 & 0.14 & 0.27 & 0.82 & 0.53 \\
    5 & Vassilis Sioros & 0.34 & 0.14 & 0.27 & 0.82 & 0.53 \\
    \midrule
    % 6 & Qwen3 VL 8b (baseline) & 0.11 & 0.03 & 0.08 & 0.30 & 0.19 \\
    6 & Intern VL 3.5 8b~\cite{wang2025internvl3_5} (baseline) & 0.26 & 0.09 & 0.2 & 0.79 & 0.48 \\
 \\
    \bottomrule
    \end{tabular}
\caption{Task 3 Summarization scores for the best baseline and top-5 teams}
\label{tab:leaderboard_summarization}
\end{table}

\subsubsection{Tasks 4: Visual Question-Answering Results}
\label{sec:results_vqa}
The approaches used by the top-5 submissions are presented below and the Table~\ref{tab:leaderboard_vqa} shows the task leaderboard.

\begin{enumerate}
    \item \textbf{DeepVitminC} \\
    This team employs a rule-based strategy to extract figure captions and surrounding document text, which are incorporated as complementary contextual signals via retrieval-augmented generation. Both caption-based and full-text contexts are used to construct enriched prompts, thereby improving semantic grounding. The resulting multimodal inputs are then used to train a vision-language model through supervised fine-tuning, yielding a Consistent Alignment Module that ensures outputs adhere to task-specific formats and semantic expectations.

    \item \textbf{Ricoh\_SRCB} \\
    The team adopts a straightforward two-stage training pipeline consisting of supervised fine-tuning followed by GRPO-based alignment. However, limited gains are observed due to suboptimal initialization and insufficiently robust data distribution during the second stage. While SFT leverages the full dataset, the GRPO phase operates on less diverse samples, resulting in marginal improvements.
    % The findings suggest that further experimentation in data curation and optimization strategies is necessary to fully exploit preference-based training.
    
    \item \textbf{DocMiner} \\
    This team's approach focuses on iterative refinement within a multi-agent framework, where each agent specializes in a specific question type. Performance is progressively improved through systematic error analysis of development data, guiding adjustments to prompt design, context structuring, and response-generation logic. This continuous optimization process enhances both the accuracy and robustness of the overall VQA system by improving the competence of individual specialized components.
    
    % \item \textbf{NCSR Demokritos \& National and Kapodistrian University of Athens (NDNKU)} \\
    \item \textbf{Vassilis Sioros} \\
    This team fine-tunes \href{https://huggingface.co/Qwen/Qwen3.5-9B}{Qwen3.5-9B} across multiple answer types within a unified multimodal framework, incorporating cropped visual inputs, localized textual context, and precomputed intermediate representations such as summaries and tables. It employs answer-type-specific token budgets and extensive preprocessing, including normalization and format-aware post-processing, to ensure output consistency. The pipeline is structured hierarchically, treating earlier tasks (e.g., table extraction and summarization) as evidence generation stages that support downstream VQA, thereby improving reasoning accuracy and contextual grounding.

    \item \textbf{VLMinators} \\
    The team utilizes parameter-efficient fine-tuning of \href{https://huggingface.co/Qwen/Qwen2.5-VL-7B-Instruct}{Qwen2.5-VL-7B-Instruct}~\cite{qwen2.5-VL} with QLoRA~\cite{dettmers2023qlora} on a diverse set of question-answer pairs spanning multiple answer types. Prompt design explicitly encodes answer-type constraints to control output structure and length, with adaptive decoding limits. A key contribution is the sequential context chaining mechanism, which integrates outputs from upstream tasks, such as figure classification, table extraction, and summarization, into the VQA prompt. This cross-task information flow significantly enhances numerical precision and contextual reasoning, particularly for factoid and paragraph-based queries.
\end{enumerate}

\begin{table}[htb]
\centering
% \tiny
% \setlength{\tabcolsep}{0.8pt}
% \renewcommand{\arraystretch}{0.95}
\scriptsize
\setlength{\tabcolsep}{3pt}
\renewcommand{\arraystretch}{1.05}
\resizebox{\columnwidth}{!}{%
    \begin{tabular}{l l c c c c c c c c c c c}
    \toprule
    \textbf{\#} & \textbf{Team} & \multicolumn{4}{c}{\textbf{Factoid}} & \textbf{Yes/No} & \multicolumn{4}{c}{\textbf{Paragraph}} & \multicolumn{1}{c}{\textbf{List}} & \textbf{Weighted} \\
    % ~ & ~ & ~ & \textbf{Rouge-1} & \textbf{Rouge-2} & \textbf{Rouge-L} & \textbf{Exact Match} & \textbf{F1-score} & \textbf{Rouge-1} & \textbf{Rouge-2} & \textbf{Rouge-L} & \textbf{BERTscore-F1} 
    % & \textbf{Set-based-F1} \\
    ~ & ~  & \textbf{R1} & \textbf{R2} & \textbf{RL} & \textbf{EM} & \textbf{F1} & \textbf{R1} & \textbf{R2} & \textbf{RL} & \textbf{BERT-F1} 
    & \textbf{Set-based-F1} & ~ \\
    \midrule
    % 1 & PAPCIC & 0.31 & 0.49 & 0.29 & 0.46 & 0.09 & 0.92 & 0.42 & 0.19 & 0.32 & 0.85 & 0.09 \\
    % 2 & Ricoh\_SRCB & 0.27 & 0.44 & 0.21 & 0.41 & 0.07 & 0.87 & 0.35 & 0.12 & 0.25 & 0.83 & 0.08 \\
    % 3 & zoloz-realdoc & 0.27 & 0.43 & 0.24 & 0.41 & 0.09 & 0.89 & 0.33 & 0.11 & 0.23 & 0.78 & 0.08 \\
    % 4 & vsioros & 0.26 & 0.44 & 0.24 & 0.40 & 0.06 & 0.86 & 0.35 & 0.14 & 0.26 & 0.77 & 0.07 \\
    % 5 & kabdelma & 0.25 & 0.40 & 0.18 & 0.37 & 0.04 & 0.85 & 0.33 & 0.10 & 0.23 & 0.82 & 0.05 \\
    1 & DeepVitminC & 0.49 & 0.29 & 0.46 & 0.09 & 0.92 & 0.42 & 0.19 & 0.32 & 0.85 & 0.09 & 0.31 \\
    2 & Ricoh\_SRCB & 0.44 & 0.21 & 0.41 & 0.07 & 0.87 & 0.35 & 0.12 & 0.25 & 0.83 & 0.08 & 0.27 \\
    3 & DocMiner & 0.43 & 0.24 & 0.41 & 0.09 & 0.89 & 0.33 & 0.11 & 0.23 & 0.78 & 0.08 & 0.27 \\
    % 4 & NDNKU & 0.44 & 0.24 & 0.40 & 0.06 & 0.86 & 0.35 & 0.14 & 0.26 & 0.77 & 0.07 & 0.26 \\
    4 & Vassilis Sioros & 0.44 & 0.24 & 0.40 & 0.06 & 0.86 & 0.35 & 0.14 & 0.26 & 0.77 & 0.07 & 0.26 \\
    5 & VLMinators & 0.40 & 0.18 & 0.37 & 0.04 & 0.85 & 0.33 & 0.10 & 0.23 & 0.82 & 0.05 & 0.25 \\
    \midrule
    % 6 & Qwen3 VL 8b (baseline) & 0.27 & 0.10 & 0.26 & 0.04 & 0.65 & 0.29 & 0.09 & 0.19 & 0.73 & 0.04 & 0.32 \\
    6 & Molmo2 8b~\cite{clark2026molmo2} (baseline) & 0.2 & 0.08 & 0.19 & 0.03 & 
    0.78 & 
    0.26 & 0.08 & 0.18 & 0.73 & 
    0.05 & 
    0.2 \\
    \bottomrule
    \end{tabular}%
}
\caption{Task 3 VQA scores for the best baseline and top-5 teams}
\label{tab:leaderboard_vqa}
\end{table}

\subsection{Discussion}
\label{sec:discussion}
In this section, we present a comparative analysis of the approaches adopted by participating teams across all competition tasks, highlighting key insights and methodological trends. We further identify the models and strategies that demonstrate consistently strong performance across tasks, providing an overview of the most effective approaches.

\subsubsection{Task 1: Classification}
\label{sec:results_discussion_classification}
The leading approach by Ricoh\_SRCB adopts a multi-stage, multi-granularity framework that decomposes the classification problem into hierarchical subtasks, complemented by dual-image inference (global and cropped views), yielding the strongest performance. In contrast, IIT\_Patna\_CV\_1 reformulates classification as a generative task using \href{https://huggingface.co/Qwen/Qwen2.5-VL-7B-Instruct}{Qwen2.5-VL-7B-Instruct}~\cite{qwen2.5-VL}, leveraging prompt design and test-time augmentation for robustness. DocMiner and VLMinators emphasize prompt engineering, contextual enrichment, and selective label hints, achieving competitive results. Overall, explicit problem decomposition and handling of class ambiguity emerge as key factors for high performance.

\subsubsection{Task 2: Data Extraction}
\label{sec:results_discussion_data_extraction}
TeleOCR-VL demonstrates the strongest performance through a comprehensive pipeline integrating structure-aware training, synthetic data augmentation, and ensemble-based inference, highlighting the effectiveness of data-centric and system-level optimization. VLMinators achieves competitive results via lightweight fine-tuning and prompt-based context injection, underscoring the strength of pretrained models. Ricoh\_SRCB and Vassilis Sioros focus on input representation and architectural enhancements, while DocMiner explores modular reasoning. The results indicate that combining data design, contextual input engineering, and model ensembling yields the most robust performance.

\subsubsection{Task 3: Summarization}
\label{sec:results_discussion_csummarization}
Top approaches reveal a transition from modular pipelines to advanced training and inference strategies. DeepVitminC employs a structured pipeline with retrieval-augmented context and alignment modules, ensuring coherent outputs. Ricoh\_SRCB enhances performance through alignment techniques such as DPO~\cite{rafailov_direct_2023} with hard negative sampling, while TeleOCR-VL focuses on inference-time robustness via multi-candidate generation and consensus-based re-ranking. Vassilis Sioros demonstrates that strong base models with efficient preprocessing and normalization can remain highly competitive. Key insights highlight the importance of contextual grounding, alignment strategies, and inference-time validation.

\subsubsection{Task 4: Visual Question-Answering}
\label{sec:results_discussion_vqa}
The VQA task exhibits a balance between pipeline engineering and training optimization. DeepVitminC leverages structured context extraction with retrieval mechanisms for strong grounding, while Ricoh\_SRCB shows limited gains from purely training-based improvements. DocMiner emphasizes iterative prompt refinement and agent-based reasoning, improving adaptability. Vassilis Sioros and VLMinators achieve strong results through sophisticated pipeline design, including answer-type-aware optimization and sequential context chaining. The findings indicate that structured context integration and task-specific optimization are critical for VQA performance.

\subsubsection{Comparative Analysis and Key Takeaways}
\label{sec:results_discussion_performance}
% Analysis of the top-performing submissions across all tasks indicates a consistent reliance on models from the Qwen family, including \href{https://huggingface.co/Qwen/Qwen2.5-VL-7B-Instruct}{Qwen2.5-VL-7B-Instruct}~\cite{qwen2.5-VL} and \href{https://huggingface.co/Qwen/Qwen3.5-9B}{Qwen3.5-9B}~\cite{qwen3.5}. While several teams also experimented with alternative vision-language models such as \href{https://deepmind.google/models/model-cards/gemini-3-1-pro/}{Gemini 3.1 Pro}, \href{https://huggingface.co/meta-llama/Llama-3.2-11B-Vision-Instruct}{Llama-3.2-11B-Vision-Instruct}, \href{https://huggingface.co/llava-hf/llava-v1.6-mistral-7b-hf}{llava-v1.6-mistral-7b-hf}~\cite{liu2023improved}, and \href{https://huggingface.co/microsoft/Phi-3.5-vision-instruct}{Phi-3.5-vision-instruct}, the strongest empirical results were consistently achieved by Qwen-based models. This trend suggests that the Qwen family currently offers superior performance and adaptability for multimodal scientific comprehension tasks.

Across all tasks, a consistent pattern emerges: top-performing approaches combine strong vision-language backbones (predominantly Qwen-based models such as \href{https://huggingface.co/Qwen/Qwen2.5-VL-7B-Instruct}{Qwen2.5-VL-7B-Instruct} and \href{https://huggingface.co/Qwen/Qwen3.5-9B}{Qwen3.5-9B} with carefully designed pipelines that integrate context, structure, and task-specific constraints. Task 1 favors explicit decomposition and ambiguity handling, while Task 2 highlights the importance of data-centric design and ensembling. Task 3 demonstrates the effectiveness of combining contextual grounding with alignment and inference-time validation, whereas Task 4 underscores the necessity of structured reasoning pipelines and context chaining. Overall, the most successful strategies are holistic---integrating data engineering, prompt design, model adaptation, and inference-time optimization, suggesting that future improvements will likely arise from unified frameworks that jointly optimize these components rather than relying on isolated advancements.

% ---- CONCLUSIONS ----
\section{Conclusions}
\label{sec:conclusions}

The ICDAR Sci-ImageMiner competition represents the first 
benchmark dataset and competition 
dedicated to scientific comprehension and reasoning over ALD/E figures.
% within the domain of materials science. 
% A benchmark dataset was 
% carefully 
The dataset is
% was curated and 
annotated by domain experts to reflect the complexity and specificity of ALD/E scientific content. 
The competition attracted substantial global participation across all four task categories, indicating strong interest from both academia and industry.
Notably, participating teams proposed 
% a diverse range of 
novel methodologies and approaches.
% technical innovations. 
Despite these advances, the results reveal that significant challenges remain, particularly in data extraction and visual question-answering, which require robust multimodal reasoning and precise scientific understanding.
These findings underscore the inherent difficulty of reliable comprehension in specialized scientific documents and highlight important directions for future research.
% The Sci-ImageMiner benchmark dataset and evaluation protocols developed for the competition will be publicly released following the event to support continued progress in this area.

% ---- ACKNOWLEDGEMENTS ----
\section{Acknowledgments}
\label{sec:acknowledgments}
We would like to thank all participants for their 
% exceptional
% efforts and 
contributions 
% to this competition, 
and the competition chairs for the opportunity to host this competition in ICDAR 2026. 
We would also like to thank the co-organizing team, Eleni Poupaki (TUE, NL), Alex Watkins (UOW, UK), Bora Karasulu (UOW, UK), and Erwin Kessels (TUE, NL). Sci-ImageMiner benchmark dataset is funded by the \href{https://www.nfdi4datascience.de/}{NFDI4DataScience} initiative (DFG, Grant ID: 460234259).
\bibliographystyle{splncs04}
\bibliography{main}

\end{document}